\documentclass[conference]{IEEEtran}
\IEEEoverridecommandlockouts
\usepackage{cite}
\usepackage{array}
\usepackage[table]{xcolor}
\usepackage{booktabs}
\usepackage{comment}
\usepackage{amsthm}
\usepackage{amsmath,amssymb,amsfonts}
\usepackage{graphicx}
\usepackage{textcomp}
\usepackage{xcolor}
\usepackage{bm}
\usepackage{subfig}
\usepackage{algorithm}
\usepackage{algorithmic}

\newtheorem{property}{Property}

\newenvironment{shrinkeq}[1]
{ \bgroup
\addtolength\abovedisplayshortskip{#1}
\addtolength\abovedisplayskip{#1}
\addtolength\belowdisplayshortskip{#1}
\addtolength\belowdisplayskip{#1}}
{\egroup\ignorespacesafterend}

\def\BibTeX{{\rm B\kern-.05em{\sc i\kern-.025em b}\kern-.08em
    T\kern-.1667em\lower.7ex\hbox{E}\kern-.125emX}}
\begin{document}

\title{Time-Varying Graph Learning  Under Structured Temporal Priors
}

\author{\IEEEauthorblockN{Xiang Zhang}
\IEEEauthorblockA{\textit{School   of   Information   Science   and   Engineering} \\
\textit{Southeast   University}\\
Nanjing, China \\
xiangzhang369@seu.edu.cn}
\and
\IEEEauthorblockN{Qiao Wang}
\IEEEauthorblockA{\textit{School   of   Information   Science   and   Engineering} \\
\textit{Southeast   University}\\
Nanjing, China \\
qiaowang@seu.edu.cn}

}

\maketitle

\begin{abstract}
This paper endeavors to learn time-varying graphs by using structured  temporal priors that assume underlying relations  between  arbitrary two graphs in the graph sequence. Different from many existing methods that only describe variations between two consecutive graphs, we propose a structure named \emph{temporal graph} to characterize the  underlying real temporal relations. Under this framework, classic priors like temporal homogeneity  is actually a special case of our temporal graph. To address computational issue, we further develop a distributed algorithm based on Alternating Direction Method of Multipliers (ADMM)  to solve the induced optimization problem. Numerical experiments on synthetic and real data demonstrate the superiorities of our method.

\end{abstract}


\section{Introduction}
\label{sec:introduction}
Inferring the  topology from data containing (hidden) structure,  which is  also  called graph learning \cite{mateos2019connecting,kalofolias2016learn, dong2016learning,dong2019learning}, has become a hot research topic  since that  prior graphs are usually unavailable for graph-based models  in many applications, e.g., graph neural networks \cite{wu2020comprehensive}. In parallel with statistical  models \cite{friedman2008sparse, yuan2007model},   graph  signal  processing (GSP)   \cite{ortega2018graph} also plays a pivotal role in graph learning, which attempts to learn graphs from  perspective  of  signal  processing.  One notable  assumption that GSP based models leverage is smoothness, under which signal values of two connected vertices with large edge  weights  tend  to  be  similar \cite{dong2016learning}. 
On the other hand, a typical feature existing in most models is that the environment is  assumed to be static such that one can learn merely a single graph from all observed data. However,  relationships   between   entities are usually time-varying  in real world. Therefore, learning a series of time-varying graphs with timestamps is a reasonable choice.

Current time-varying graph learning methods attempt to jointly learn graphs of all time slots by exploiting prior assumptions about evolutionary patterns of dynamic graphs \cite{kalofolias2017learning}.  One may find that  the most used assumptions here is temporal homogeneity  \cite{yamada2019time}, under which only a small number of edges  are allowed to change between two consecutive graphs. The essence of prior assumptions like temporal homogeneity is to establish temporal relations  between  graphs  of  different  time slots using prior knowledge, which  are crucial for learning time-varying graphs since they actually bring structural information, in addition to data,  to learning process.

Albeit interesting, assumption of temporal homogeneity  only cares about variations between graphs in neighboring time slots and treat them equally. 
Obviously, this assumption is simple enough but it  may be inconsistent with the real temporal relations in some applications. Here we take crowd flow networks of urban area as an example. The variations of crowd flow networks at different time periods in a day are not uniform due to the difference of travel behaviour \cite{thanou2017learning}. For example, the patterns of networks in early morning (1 a.m.--5 a.m.) are apparently different from those  in rush hours  (7 a.m.--9 a.m.). Thus, it is not reasonable to treat all 
variations equally. Furthermore, common knowledge tells us that networks in the same time period of two different working days, e.g., 10 a.m. in Monday and Tuesday, are also  similar. Capturing this periodic pattern is beyond the ability of the temporal homogeneity assumption. 

To this end, a more general time-varying graph learning method  should be proposed by generalizing the assumption of temporal homogeneity. In this paper,  a  flexible  structure named  \emph{temporal graph} is leveraged to describe structured  temporal  relations of time-varying graphs. In temporal graph,  relations  between  graphs  of  any  paired  time  slots, not limited to adjacent time slots,  can  be  established, and we use  weights  to  measure the  ”closeness”  of these  relations.  Therefore, temporal homogeneity assumption \cite{yamada2019time} is a special case of our framework. Furthermore, the algorithm for solving the classic model \cite{yamada2019time} suffers from increasing complexity as the number of time slots since it learns  graphs of all time slots jointly. To address computational issue, a distributed algorithm based on  Alternating Direction Method of Multipliers (ADMM) is developed to solve the induced optimization problem, which can save  considerable time when the number of time periods is large. Numerical tests illustrate that our method outperforms the state-of-art methods in face of intricate temporal structures.

\section{Preliminaries}
\label{sec:preliminaries}
We  will learn undirected  graphs $\mathcal{G}$ with non-negative  weights.  Given $N$ observed signals $\mathbf{x}_1,\ldots,\mathbf{x}_N \in \mathbb{R}^d$  generated from $\mathcal{G}$, graph learning is aimed to infer the adjacency matrix $\mathbf{A}\in \mathbb{R}^{d\times d}$ of $\mathcal{G}$. Under smoothness priors, it is equivalent to solving the following problem \cite{kalofolias2016learn}
\begin{shrinkeq}{-0.8ex}
\begin{align}
    \underset{\mathbf{A}\in \mathcal{A}}{\mathrm{min}}\; \lVert \mathbf{A} \circ \mathbf{R} \rVert_1   -\alpha \mathbf{1}^{\top}\mathrm{log}(\mathbf{A}\mathbf{1}) + \frac{\beta}{2}\lVert \mathbf{A}\rVert_{\mathrm{F}}^2,
    \label{GL-1}    
\end{align}
\end{shrinkeq}
where $\circ$ is Hadamard product and $\mathbf{1} = [1,\ldots,1]^{\top}\in \mathbb{R}^{d}$ is a column vector of ones. Parameters $\alpha$ and $\beta$ are predefined constants. Furthermore, $\mathcal{A}$ is the  set defined as \cite{kalofolias2016learn} 
\begin{shrinkeq}{-0.8ex}
\begin{align}
        \mathcal{A} = \left\{ \mathbf{A} : \mathbf{A}\in \mathbb{R}^{d\times d}_+,  \mathbf{A} = \mathbf{A}^{\top},\mathrm{diag}(\mathbf{A}) = \mathbf{0}\right\},
    \label{GL-2}
\end{align}
\end{shrinkeq}
where $\mathbb{R}_+$ is the set of nonnegative real numbers and $\mathbf{0} \in \mathbb{R}^d$ is a  column vector of zeros. For data matrix $\mathbf{X} \in \mathbb{R}^{d\times N} = [\mathbf{x}_1,\ldots,\mathbf{x}_N] = [\tilde{\mathbf{x}}_1,\ldots,\tilde{\mathbf{x}}_d]^{\top}$, the  pairwise distance matrix $\mathbf{R}\in \mathbb{R}^{d\times d}$ in \eqref{GL-1} is defined as
\begin{shrinkeq}{-0.8ex}
\begin{align}
        \mathbf{R}_{[ij]} = \lVert \tilde{\mathbf{x}}_i - \tilde{\mathbf{x}}_j\rVert_2^2,
    \label{GL-3}   
\end{align}
\end{shrinkeq}
where $\mathbf{R}_{[ij]}$ is the $(i,j)$ entry of $\mathbf{R}$. The first term of \eqref{GL-1} is the smoothness of the observed signals over $\mathcal{G}$. Besides, the  second  and third term control  degrees  of  each node  and sparsity  of  edges \cite{kalofolias2016learn}. Note that $\mathbf{A}$ is a symmetric matrix with diagonal entries equal to zero, and hence the number of free variables of $\mathbf{A}$ is  $p\triangleq\frac{d(d-1)}{2}$. We define a vector $\mathbf{w}\in \mathbb{R}^p$ whose entries are the upper right variables of $\mathbf{A}$. Therefore, problem \eqref{GL-1} can be rewritten as \cite{kalofolias2016learn}
\begin{shrinkeq}{-0.8ex}
\begin{align}
       \underset{\mathbf{w}\geq 0}{\mathrm{min}} \,\, f(\mathbf{w}) =  \underset{\mathbf{w}\geq 0}{\mathrm{min}}\,\, 2\mathbf{r}^{\top}\mathbf{w} -\alpha\mathbf{1}^{\top}\mathrm{log}(\mathbf{S}\mathbf{w}) + \beta\lVert \mathbf{w} \rVert_2^2,
    \label{reformulation-of-GL-W}
\end{align}
\end{shrinkeq}
where the linear operator $\mathbf{S}$  satisfies $\mathbf{S}\mathbf{w} = \mathbf{A}\mathbf{1}$  and $\mathbf{r}$ is the vector form of the upper right variables of $\mathbf{R}$.

Under these notations, the time-varying graph learning will produce  a series of graphs $\mathbf{w}_1,\ldots,\mathbf{w}_T$ using signals $\mathbf{X}_1,\mathbf{X}_2,..\mathbf{X}_T$ collected during $T$ time periods, where $\mathbf{X}_t \in \mathbb{R}^{d\times N}$ is the data matrix of time slot $t$.  Specifically,  temporal homogeneity  assumption  based model is formulated as \cite{yamada2019time}
\begin{shrinkeq}{-1ex}
\begin{align}
    &\underset{\mathbf{w}_t\geq 0}{\mathrm{min}}\,\, \sum_{t=1}^T f_t(\mathbf{w}_t) + \eta \sum_{t=2}^T \left\lVert\mathbf{w}_{t} - \mathbf{w}_{t-1}\right\rVert_1\notag\\[-0.2ex]
    =\,\,&\underset{\mathbf{w}_t\geq 0}{\mathrm{min}}\,\,
     \sum_{t=1}^T 2\mathbf{r}_t^{\top}\mathbf{w}_t -\alpha\mathbf{1}^{\top}\mathrm{log}(\mathbf{S}\mathbf{w}_t) + \beta\lVert \mathbf{w}_t \rVert_2^2 \notag\\[-1.5ex]
     &\quad\quad +\eta \sum_{t=2}^T \left\lVert \mathbf{w}_{t} - \mathbf{w}_{t-1}\right\rVert_1,
    \label{classic time varying graph learning}
\end{align}
\end{shrinkeq}
where $\eta$ is a global parameter controls the weight of temporal priors and $\mathbf{r}_t$ is calculated using $\mathbf{X}_t$. The last term of \eqref{classic time varying graph learning} indicates that only a small number of edges  are allowed to change between two consecutive graphs.

\section{Proposed Framework}
\label{sec:Proposed Framework}
\begin{figure}[t]
    \centering
       \includegraphics[width=0.7\linewidth]{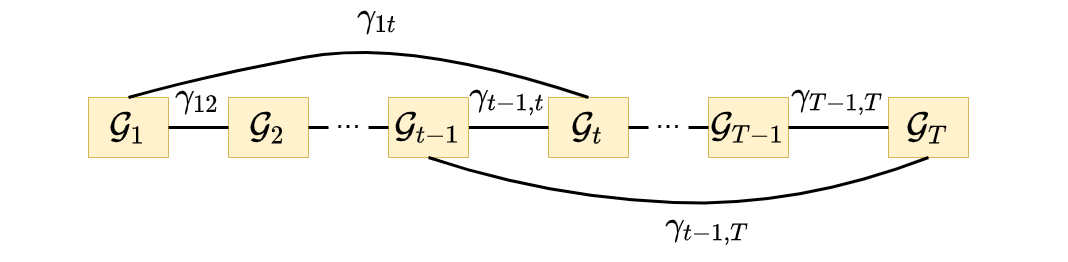}
    	\caption{A prototype of temporal  graph  structure}
	    \label{fig-prototypeModel}
	    \vspace{-2em}
\end{figure}

Observe \eqref{classic time varying graph learning} and we can find that temporal homogeneity prior only imposes constraints on variations of graphs between two adjacent time slots equally, which is too simple and may fail to characterize the  temporal relations in real world. In this paper, we suggest a general structure named \emph{temporal graph} to describe temporal relations of time-varying graphs. The temporal graph $\mathcal{G}_N$ is a graph structure whose nodes represent graphs of $T$ time slots and edges indicate the relationships between the connected nodes, i.e, constraints on variations between the corresponding graphs. As shown in Fig.\ref{fig-prototypeModel}, temporal graph is undirected but with nonnegative weighted edges. Any two nodes can be connected in temporal graph, e.g., $\gamma_{1t}$ in Fig.\ref{fig-prototypeModel}, instead of  allowing merely two consecutive graphs connection. Furthermore, we give up treating these temporal constraints equally and use edge weights to measure the ``importance" of them. Clearly, temporal graph is  more general and able to describe intricate temporal structures  in real world.

\label{sec:temporal graph}

Formally, we define a  temporal graph $\mathcal{G}_N =\{\mathcal{V}_N,\mathcal{E}_N\}$, where $\mathcal{V}_N$ is the node set containing graphs of all time slots and $\mathcal{E}_N$ is the edge set containing connections between these graphs. In this paper, we suppose that  there are $T$ nodes and $s$ edges in $\mathcal{G}_N$, i.e., $\left|\mathcal{V}_N\right| = T$ and $\left|\mathcal{E}_N\right| = s$. Time-varying graph learning using temporal graph  is formulated as
\begin{shrinkeq}{-0.3ex}
\begin{align}
    \underset{\mathbf{w}_t\geq 0 }{\mathrm{min}}\,\, \sum_{t\in \mathcal{V}_N}\,f_t(\mathbf{w}_t)+ \eta \sum_{(i,j) \in \mathcal{E}_N} \gamma_{ij} \left\lVert\mathbf{w}_i - \mathbf{w}_{j}\right\rVert_1,
    \label{TSGL-basic formulation}
\end{align}
\end{shrinkeq}
where $\gamma_{ij}$ is the relative weights
between the $i$-th and  $j$-th time slots. Parameter $\eta$ is used to scale the weight of edge objectives relative to node objectives. Note that, \eqref{TSGL-basic formulation}  will boil down to \eqref{classic time varying graph learning} if we build $\mathcal{G}_N$ as a chain with equal weights.

We should mention that the design of $\mathcal{G}_N$, which is obtained by our prior knowledge of temporal relations, is critical for the learning performance. Better knowledge of  temporal relations in real world does help improve the performance. The main contribution of temporal graph $\mathcal{G}_N$ is providing a more flexible structure to describe temporal relations fusing our prior knowledge. In the worst case, when we have no structured temporal priors, our framework can still boil down to model \eqref{classic time varying graph learning} but provides a more efficient algorithm introduced in the  next section.

\section{ADMM based algorithm}
\label{sec:algorithm}
The algorithm for solving \eqref{classic time varying graph learning} attempts to learn $\mathcal{G}_1,...,\mathcal{G}_T $ in a centralized fashion, resulting in increasing complexity as $T$. We hence develop a novel distributed algorithm based on ADMM framework \cite{nocedal2006numerical, boyd2004convex, boyd2011distributed} to solve \eqref{TSGL-basic formulation}. The algorithm is able to learn graphs of different time slots in parallel, showing its efficiency when $T$ is large. For an edge $(i,j)\in \mathcal{E}_N$, we first introduce a consensus variable of $\mathbf{w}_i$, denoted as $\mathbf{z}_{ij}$. In fact, $\mathbf{z}_{ij}$ represents the connection starting from $i$ to $j$. For the same edge, $\mathbf{z}_{ji}$ is the consensus variable of $\mathbf{w}_j$. With consensus variables,  \eqref{TSGL-basic formulation} is equivalent to the following problem 
\begin{shrinkeq}{-0.5ex}
\begin{align}
      &\underset{\mathbf{w}_t\geq 0 }{\text{min}}\,\, \sum_{t\in \mathcal{V}_N}\,f_t(\mathbf{w}_t) +\eta \sum_{(i,j)\in \mathcal{E}_N}\gamma_{ij}\left\lVert\mathbf{z}_{ij} -\mathbf{z}_{ji}\right\rVert_1\notag\\[-0.5ex]
        &\text{s.t.}\,\,\mathbf{w}_i= \mathbf{z}_{ij},\,\, \text{for}\,\, i = 1,...,T \,\,\mathrm{and}\,\, j\in \mathcal{N}(i), 
        \label{reformulation-1}
\end{align}
\end{shrinkeq}
where $\mathcal{N}(i)$ denotes the set of all the nodes that are connected with node $i$ in $\mathcal{G}_N$. We define a matrix $\mathbf{W}\in \mathbb{R}^{p\times T} \triangleq  [\mathbf{w}_1,...,\mathbf{w}_T]$ containing all primal variables. In addition, matrices of consensus variables $\mathbf{Z}\in\mathbb{R}^{p\times 2s}$ and dual variables $\mathbf{U}\in \mathbb{R}^{p\times 2s} $ are also defined. For the $n$-th edge $(i,j)$ in temporal graph $\mathcal{G}_N$, $n=1,\ldots,s$, the corresponding consensus variable vectors $\mathbf{z}_{ij},\mathbf{z}_{ji}$  are the $(2n-1)$-th and  $2n$-th columns of $\mathbf{Z}$, respectively. This also holds true for matrix $\mathbf{U}$.
The scaled form of augmented Lagrangian of \eqref{reformulation-1} is  obtained as
\begin{shrinkeq}{-0.5ex}
\begin{align}
    &L_{\rho}(\mathbf{W},\mathbf{Z},\mathbf{U}) \notag\\[-0.7ex]
    =&
    \sum_{t\in \mathcal{V}_N}\,f_t(\mathbf{w}_t) +\eta \sum_{(i,j)\in \mathcal{E}_N}\gamma_{ij}\left\lVert\mathbf{z}_{ij} -\mathbf{z}_{ji}\right\rVert_1  \notag\\[-0.7ex]
    & + \sum_{(i,j)\in \mathcal{E}_N} \bigg{(} \frac{\rho}{2}\left(\lVert\mathbf{u}_{ij}\rVert_2^2+\lVert\mathbf{u}_{ji}\rVert_2^2 \right)  \notag\\[-0.7ex]
    & +\frac{\rho}{2}\left(\lVert\mathbf{w}_i - \mathbf{z}_{ij} +\mathbf{u}_{ij} \rVert_2^2 +\lVert\mathbf{w}_j - \mathbf{z}_{ji} +\mathbf{u}_{ji} \rVert_2^2\right)\bigg{)},
    \label{lagrangian-form2}
\end{align}
\end{shrinkeq}
where $\rho>0$ is an ADMM penalty parameter \cite{boyd2011distributed}. Following ADMM framework, we alternately update $\mathbf{W},\mathbf{U}$ and $\mathbf{Z}$.

\emph{1)} {Update $\mathbf{W}$}:
For $\mathbf{W} = [\mathbf{w}_1,\ldots,\mathbf{w}_T]$, the update of $\left (\mathbf{w}^{k+1}_t, 1\leq t \leq T\right)$  is as follows
\begin{shrinkeq}{-0ex}
\begin{align}
      &\left (\mathbf{w}^{k+1}_t,1\leq t \leq T \right) \notag\\[-0.5ex]
       =& \underset{\mathbf{w}_t>0}{\text{argmin}}\,\,     \sum_{t\in\mathcal{V}_N}\,f_t(\mathbf{w}_t) \notag\\[-0.5ex]
       & +\frac{\rho}{2}\sum_{(i,j)\in \mathcal{E}_N}\left(\lVert\mathbf{w}_i - \mathbf{z}^k_{ij} +\mathbf{u}^k_{ij} \rVert_2^2 +\lVert\mathbf{w}_j - \mathbf{z}^k_{ji} +\mathbf{u}^k_{ji} \rVert_2^2\right).
\label{Update-w-full}
\end{align}
\end{shrinkeq}
Obviously, we can update each $\mathbf{w}^{k+1}_t$ separately,
\begin{shrinkeq}{-1ex}
    \begin{align}
      \mathbf{w}^{k+1}_t
       = \underset{\mathbf{w}_t>0}{\text{argmin}}\,\,     f_t(\mathbf{w}_t) +\frac{\rho}{2}\sum_{j\in \mathcal{N}(t)}\lVert\mathbf{w}_t - \mathbf{z}^k_{tj} +\mathbf{u}^k_{tj} \rVert_2^2.
    \label{Update-w-t}
    \end{align}
\end{shrinkeq}
If we let $\bm{\theta}_t^{k}\triangleq \frac{1}{m}\sum_{j\in \mathcal{N}(t)} \left(\mathbf{z}_{tj}^k - \mathbf{u}_{tj}^k \right)$, where $m = |\mathcal{N}(t)|$,  \eqref{Update-w-t} can be reformulated as
\begin{shrinkeq}{-1ex}
      \begin{align}
       \mathbf{w}^{k+1}_t= &\underset{\mathbf{w}_t>0}{\text{argmin}}\,\,f_t(\mathbf{w}_t) + \frac{m\rho}{2}\lVert \mathbf{w}_t -  \bm{\theta}_t^{k} \rVert_2^2 \notag\\[-0.5ex]
       \triangleq\, &\underset{\mathbf{w}_t>0}{\text{argmin}}\,\, g_t(\mathbf{w}_t).
    \label{reformulate-update-w-t}  
    \end{align}
\end{shrinkeq}
In this paper, we use projected gradient descent (PGD) algorithm \cite{calamai1987projected} to solve problem \eqref{reformulate-update-w-t}. The gradient of the objective function of \eqref{reformulate-update-w-t} is as follows
\begin{shrinkeq}{-1ex}
    \begin{align}
       \nabla g_t(\mathbf{w}_t) = 2\mathbf{r}_t + 2\beta\mathbf{w}_t + m\rho(\mathbf{w}_t -  \bm{\theta}_t^{k}) - \alpha \mathbf{S}^{\top}\left(\mathbf{S}\mathbf{w}_t\right)^{.(-1)},
    \end{align}
\label{gradient} 
\end{shrinkeq}
where $.(-1)$ is an elementwise reciprocal operator.  We  set $\mathbf{y}^0 = \mathbf{w}_t^k$ and iteratively update $\mathbf y^r$ using 
\begin{shrinkeq}{-1ex}
    \begin{align}
        \mathbf{y}^{r+1} = \left(\mathbf{y}^{r} - \epsilon \nabla g_t(\mathbf{y}^r)\right )_+,
    \label{update-w-pgd}
    \end{align}
\end{shrinkeq}
until it converges to $\mathbf{y}^*$ with a certain precision, where $(\cdot)_+ \triangleq \text{max}(\cdot,0)$, $r$ is the number of iterations of PGD algorithm and $\epsilon$  the step size. After obtaining  the solution $\mathbf{y}^*$ of \eqref{reformulate-update-w-t}, we  set $\mathbf{w}_{t}^{k+1} = \mathbf{y}^*$. Note that all $\mathbf{w}_t$ can be updated in parallel.

\emph{2)} {Update $\mathbf{Z}$}: For each edge $(i,j) \in \mathcal{E}_N$, we can update the corresponding column vectors $\mathbf{z}_{ij}, \mathbf{z}_{ji}$ of $\mathbf{Z}$ as follows 
\begin{shrinkeq}{-0.7ex}
    \begin{align}
       &\mathbf{z}^{k+1}_{ij},\mathbf{z}^{k+1}_{ji}\notag\\[-0.5ex] =&\underset{\mathbf{z}_{ij},\mathbf{z}_{ji}} {\text{argmin}}\,\,\eta \gamma_{ij} \left\lVert\mathbf{z}_{ij} - \mathbf{z}_{ji}\right\rVert_1\notag\\[-0.5ex]
       &+\frac{\rho}{2} \bigg{(} \left\lVert \mathbf{w}^{k+1}_i - \mathbf{z}_{ij} + \mathbf{u}^{k}_{ij}\right\rVert_2^2
       +\left\lVert \mathbf{w}^{k+1}_{j} - \mathbf{z}_{ji} + \mathbf{u}^{k}_{ji}\right\rVert_2^2 \bigg{)}.
    \label{reformulate-update-z-t}
    \end{align}
\end{shrinkeq}
It is difficult to solve \eqref{reformulate-update-z-t} due to that  $\mathbf{z}_{ij}$ and  $\mathbf{z}_{ji}$ are coupled with each other in $\left\lVert\mathbf{z}_{ij} - \mathbf{z}_{ji}\right\rVert_1$. Inspired by the method proposed in \cite{hallac2017network}, we define a function $\tilde{\psi}$
\begin{shrinkeq}{-0.7ex}
    \begin{align}
        \tilde{\psi}\left(\begin{bmatrix} \mathbf{z}_{ij}  \\ \mathbf{z}_{ji} \\ \end{bmatrix} \right) = \left\lVert\mathbf{z}_{ij} - \mathbf{z}_{ji}\right\rVert_1,
   \label{definition-psi}
    \end{align}
\end{shrinkeq}
with which \eqref{reformulate-update-z-t} can be solved by
\begin{shrinkeq}{-0.7ex}
    \begin{align}
        \begin{bmatrix} \mathbf{z}_{ij}^{k+1}  \\ \mathbf{z}_{ji}^{k+1} \\ \end{bmatrix} = \mathrm{prox}_{\frac{\eta\gamma_{ij}}{\rho} \tilde{\psi}} \left(   \begin{bmatrix} \mathbf{u}_{ij}^k +\mathbf{w}_{i}^{k+1} \\ \mathbf{u}_{ji}^k +\mathbf{w}_{j}^{k+1} \\ \end{bmatrix}\right),
    \label{reformulate-update-z}    
    \end{align}
\end{shrinkeq}
where $\mathrm{prox}_{\frac{\eta\gamma_{ij}}{\rho} \tilde{\psi}}(\cdot)$ is the proximal operator of function $\tilde{\psi}$ \cite{parikh2014proximal}. However, we have no knowledge of the closed form of the operator $\mathrm{prox}_{\frac{\eta\gamma_{ij}}{\rho} \tilde{\psi}}(\cdot)$. Hence a property of proximal operators mentioned in  \cite{hallac2017network} might be introduced here.
\begin{property}
    If a function $h_1(\mathbf{v}) = h_2(\mathbf{G}\mathbf{v}+\mathbf{H})$, and $\mathbf{G}\mathbf{G}^{\top} =\frac{1}{\lambda} \mathbf{I}$, where $\mathbf{I}$ is an identity matrix, then
    \begin{shrinkeq}{-1ex}
           \begin{align}
        &\mathrm{prox}_{h_1}(\mathbf{v}) \notag\\
        =& (\mathbf{I} - \lambda \mathbf{G}^{\top}\mathbf{G})\mathbf{v} + \lambda \mathbf{G}^{\top}(\mathrm{prox}_{\frac{1}{\lambda}h_2}(\mathbf{G}\mathbf{v} + \mathbf{H})-\mathbf{H}).
        \label{property-proximal-operator}
    \end{align}
    \end{shrinkeq}
\label{property-1}
\vspace{-1.5em}
\end{property}
In our problem,  $h_1 =\tilde{\psi}, h_2 = \ell_1$ norm, $\mathbf{G} = [-\mathbf{I} \,\,\, \mathbf{I}] $,  $\mathbf{H}$ is zero matrix and $\lambda = \frac{1}{2}$. According to Property \ref{property-1}, the following update can be easily reached for \eqref{reformulate-update-z},
\begin{equation}
    \begin{aligned}
        &\begin{bmatrix} \mathbf{z}_{ij}^{k+1}  \\ \mathbf{z}_{ji}^{k+1} \\ \end{bmatrix} = \frac{1}{2}\begin{bmatrix} \mathbf{u}_{ij}^{k} + \mathbf{w}_{i}^{k+1} + \mathbf{u}_{ji}^{k} + \mathbf{w}_{j}^{k+1}\\ \mathbf{u}_{ij}^{k} + \mathbf{w}_{i}^{k+1} + \mathbf{u}_{ji}^{k} + \mathbf{w}_{j}^{k+1} \\ \end{bmatrix} \\
        &+\frac{1}{2}\begin{bmatrix} -\text{prox}_{\frac{2\eta\gamma_{ij}}{\rho}\lVert\cdot\rVert_1}\left(
        \mathbf{w}_{j}^{k+1} + \mathbf{u}_{ji}^{k} - \mathbf{w}_{i}^{k+1}
         -\mathbf{u}_{ij}^{k} \right)  \\ \text{prox}_{\frac{2\eta\gamma_{ji}}{\rho}\lVert\cdot\rVert_1}\left(        \mathbf{w}_{j}^{k+1} + \mathbf{u}_{ji}^{k} - \mathbf{w}_{i}^{k+1}
         -\mathbf{u}_{ij}^{k} \right)  \\ \end{bmatrix}.
    \end{aligned}
    \label{rewrite-update-z}
\end{equation}
Now each column of $\mathbf{Z}$ can be updated in parallel.

\begin{shrinkeq}{-0.5ex}
        \begin{align}
       \mathbf{u}_{ij}^{k+1} &= \mathbf{u}_{ij}^{k} + \mathbf{w}_i^{k+1} - \mathbf{z}_{ij}^{k+1}\notag \\
      \mathbf{u}_{ji}^{k+1}& = \mathbf{u}_{ji}^{k} + \mathbf{w}_j^{k+1} - \mathbf{z}_{ji}^{k+1}.
     \label{update-u}   
    \end{align}
\end{shrinkeq}

In summary, our algorithm can be implemented in a distributed fashion since the columns of  $\mathbf{W},\mathbf{Z}$ and $\mathbf{U}$ can all be updated in parallel. The global convergence is also guaranteed by ADMM framework since \eqref{TSGL-basic formulation} is a convex problem. Furthermore, the 
stopping criterion is that the primal and dual residuals are  both below a given tolerance. More details can be referred in \cite{boyd2011distributed}.

\begin{algorithm}[t] 
\caption{ADMM based algorithm} 
\begin{algorithmic}[1] 
\REQUIRE ~~\\ 
$\alpha$, $\beta$, $\eta$, $\rho$, the predefined  $\mathcal{G}_N$, signals $\mathbf{X}_1$,..., $\mathbf{X}_T$ \\
\ENSURE ~~\\ 
The learned graph $\mathbf{w}_1$, \ldots ,$\mathbf{w}_T$\\
\STATE Initialize  $\mathbf{w}_t^0$,  $\mathbf{z}_{ij}^0 $ and  $\mathbf{u}_{ij}^0 $ for $t \in \mathcal{V}_N$, $(i,j) \in \mathcal{E}_N$, set $k=0$\\

\WHILE{stop criterion not satisfied}
\STATE Update $\mathbf{w}_1^{k+1}$, \ldots ,$\mathbf{w}_T^{k+1}$ using PGD in parallel
\STATE Update $\mathbf{z}_{ij}^{k+1}$, $\mathbf{z}_{ji}^{k+1}$ for $(i,j) \in \mathcal{E}_N$
using \eqref{rewrite-update-z} in parallel
\STATE Update $\mathbf{u}_{ij}^{k+1}$,$\mathbf{u}_{ji}^{k+1}$  for $(i,j) \in \mathcal{E}_N$ using \eqref{update-u} in parallel
\STATE $k = k + 1$
\ENDWHILE
\RETURN $\mathbf{w}_1^{k}$, $\mathbf{w}_2^{k}$, \ldots ,$\mathbf{w}_T^{k}$ 
\end{algorithmic}
\label{alg:TSGL} 
\end{algorithm}

\section{Numerical Experiments}
\label{sec:experiments}
\subsection{Synthetic Data}

\begin{figure}[t]
    \centering
       \includegraphics[width=0.6\linewidth]{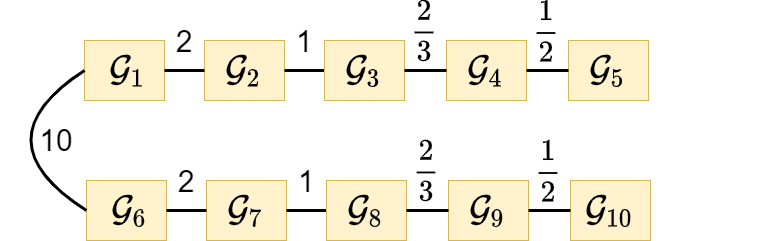}
    	\caption{Non-chain structure}
	    \label{fig-Mode1}
	    \vspace{-0.5em}
\end{figure}

We first validate the strengths of our framework and algorithm using synthetic data. The temporal structure we use is shown in Fig.\ref{fig-Mode1}. It is a unchained structure where $\mathcal{G}_6$ is connected with $\mathcal{G}_1$ instead of $\mathcal{G}_5$. To obtain time-varying graphs, an initial RBF graph $\mathcal{G}_1$ with 20 vertices is generated in the same way as  \cite{dong2016learning}. After that,  $\mathcal{G}_2$ is obtained by changing edges in $\mathcal{G}_1$ randomly and  the number of the changed edges is inverse proportion with the edge weights of $\mathcal{G}_N$ in Fig.\ref{fig-Mode1}. Following this way, we can generate other graphs sequentially. We emphasis that $\mathcal{G}_6$ is generated based on $\mathcal{G}_1$ instead of $\mathcal{G}_5$. Smooth graph signals $\mathbf{X}_t$ of each $\mathcal{G}_t$ are generated by the same way introduced in  \cite{yamada2019time}. The adopted evaluation metrics are Matthews correlation coefficient  ($\mathrm{MCC}$) \cite{powers2020evaluation} and relative error, each averaged over all time. $\mathrm{MCC}$ is a metric representing the accuracy of the estimated graph topology and its value is between -1 and 1 (-1 represents completely wrong detection while +1 means completely right detection). Relative error is defined as $\lVert \mathbf{A}^* - \mathbf{A}_{\mathrm{gt}} \rVert_{\mathrm{F}}/\lVert \mathbf{A}_{\mathrm{gt}} \rVert_{\mathrm{F}}$, where $\mathbf{A}^*$ is the learned adjacency matrix and $ \mathbf{A}_{\mathrm{gt}}$ is the groundtruth. Three baselines are leveraged, i.e.,   SGL (learn graphs of each time periods independently), TVGL-Tikhonov    \cite{kalofolias2017learning} and TVGL-Homogeneity   \cite{yamada2019time}. The last two are time-varying models with a chained temporal structure. Following the method of parameter selection in \cite{kalofolias2016learn}, we fix $\alpha = 2$ and find the best $\beta$ by grid search \cite{kalofolias2016learn}. Furthermore, we choose $\eta$ that  maximizes $\mathrm{MCC}$, which is 2.5. In ADMM framework,  $\rho$ is set to be 0.5 and tolerance values are set to be $10^{-3}$ (both relative and absolute tolerance)  \cite{boyd2011distributed}. The parameters of  baselines are all selected as the ones corresponding to the best MCC values. The following results are the average of 20 independent  experiments. All algorithms are implemented by python and run on an Intel(R) Xeon(R) CPU with 2.10GHz clock speed and 256GB of RAM. 

\begin{figure}[t] 
    \centering
	  \subfloat[]{
       \includegraphics[width=0.47\linewidth,height=0.4\linewidth]{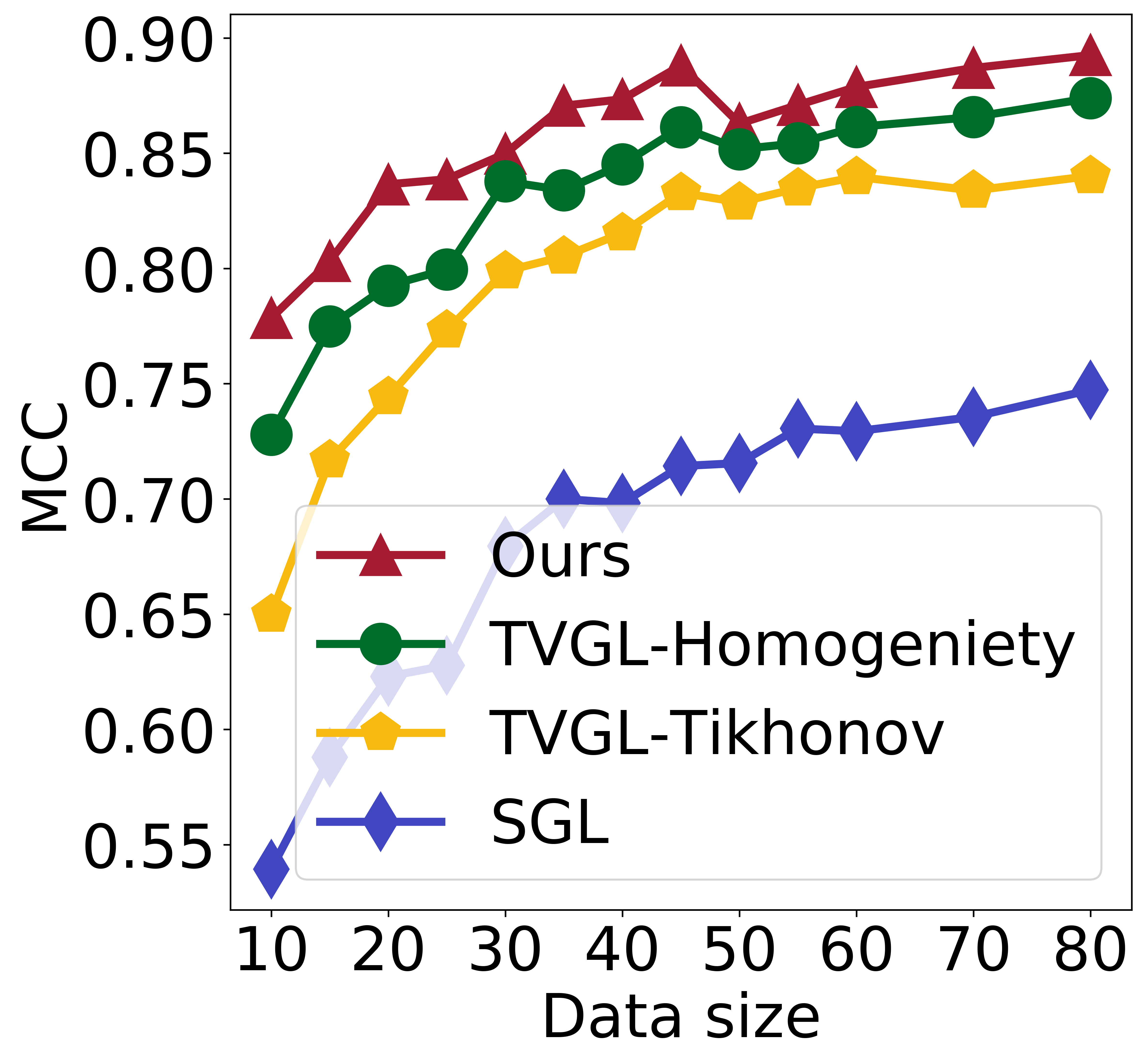}}\hspace{0.2mm}
	  \subfloat[]{
        \includegraphics[width=0.47\linewidth,height=0.4\linewidth]{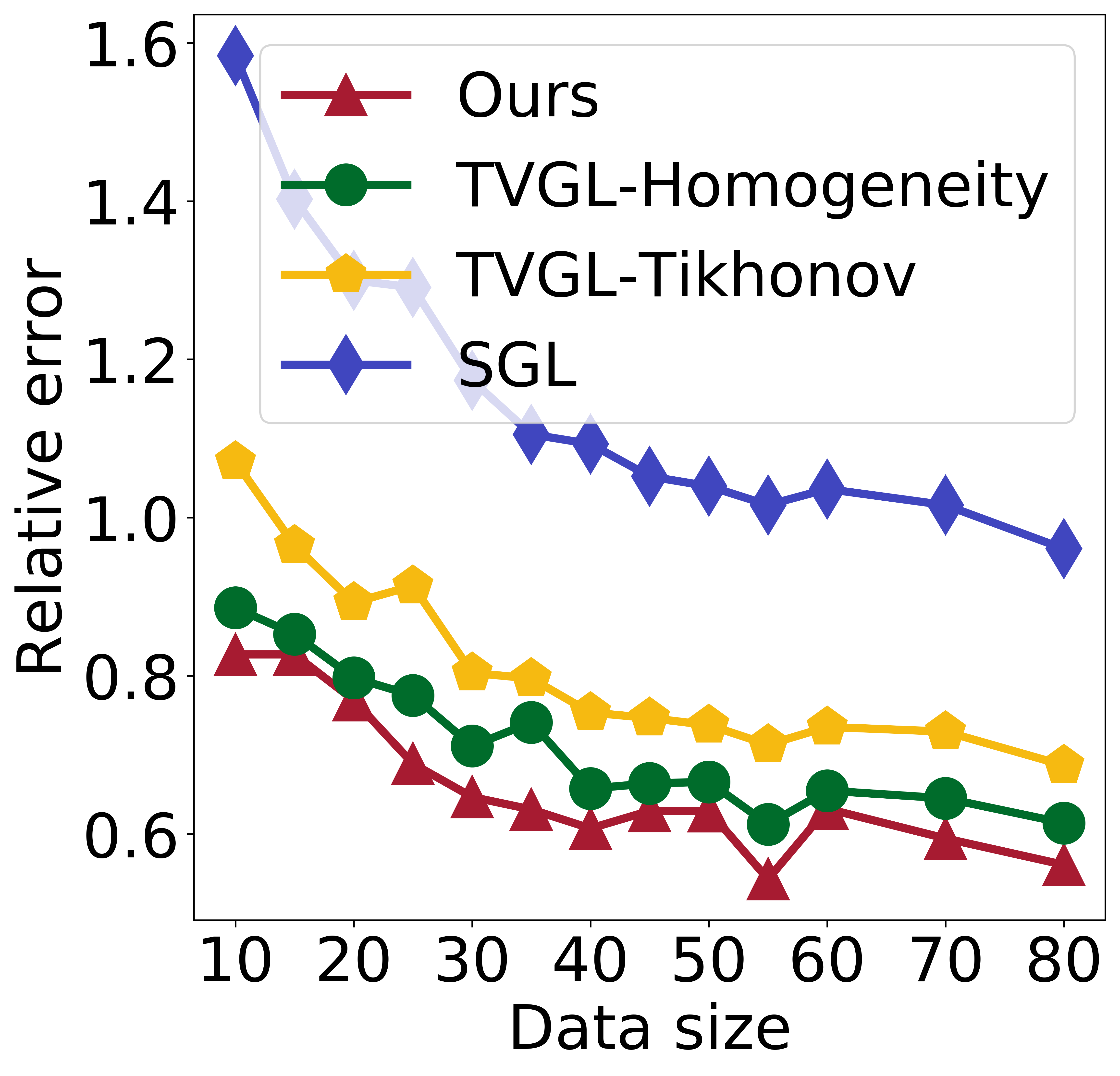}}
    	\caption{Performance of the learned graph with different sample size (a) $\mathrm{MCC}$; (b) Relative error }
    	\label{fig-mode1-performance}
    	\vspace{-0.5em}
\end{figure}

Figure \ref{fig-mode1-performance} shows the performance of different data size $N$ of each time slots. We can observe that SGL reaches the worst performance since no temporal priors are exploited. The performance of TVGL-Tikhonov and TVGL-Homogeneity is inferior to ours due to that their chain structures fail to characterize the real temporal structure depicted in Fig.\ref{fig-mode1-performance}. On the contrary, our framework is able to describe the unchained structure easily thanks to the strong representation ability of temporal graph. Therefore, our method is superior to other  models when faced with intricate temporal structures.

\begin{table}[t]
    \setlength\tabcolsep{3pt}
	\centering
	\caption{Running time (s) v.s. $T$: all results 
	take the form of logarithm ($\log_{10}$) }
	\begin{tabular}{ccccccccc}
		\toprule
		 $T$ & 2  & 5 & 10 & 15 & 20 & 25 & 30 & 35\\
		\midrule
	    \textbf{Classic} & 1.575 & 2.562 &3.231 & 3.665 & 3.901 & 4.083 & 4.187 & 4.304 \\
	    \textbf{Ours} &  0.802  &1.473 & 1.947 &2.099 & 2.136 &2.202 &2.523& 2.502\\
		\bottomrule
	\end{tabular} 
 \label{table-running time}
 \vspace{-0.8em}
\end{table}

We also compare the efficiency of our algorithm with that of classical centralized algorithm in \cite{yamada2019time}. We fix $d=100$ and apply our method to chain temporal structure problems defined in \cite{yamada2019time}. This is feasible since chain structure is a special case of our framework. We implement our algorithm in a distributed way. Our code is run on different cores of a single machine and 25 cores are used. The results of two algorithms are listed in Table \ref{table-running time}. We take logarithm ($\log_{10}$) on the results for ease of presentation. As shown in Table \ref{table-running time}, the running time of TVGL-homogeneity is significantly greater than ours especially when $T$ is large. The runtime of our algorithm increases slowly with $T$ thanks to the distributed feature. A 
great increase occurs in our algorithm when $T=30$, which is  caused by that the number of used cores is 25 and additional waiting time is required when $T>25$.

\begin{figure}[t]
    \centering
       \includegraphics[width=0.9\linewidth]{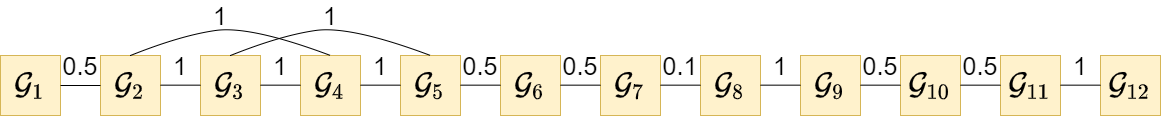}
    	\caption{The designed temporal structure}
	    \label{fig-2BE6-1}
	    \vspace{-1.8em}
\end{figure}

\begin{figure*}[t] 
    \centering
	  \subfloat[Ours: $\mathcal{G}_3$ (2 a.m.-3 a.m.)]{
       \includegraphics[width=0.18\linewidth]{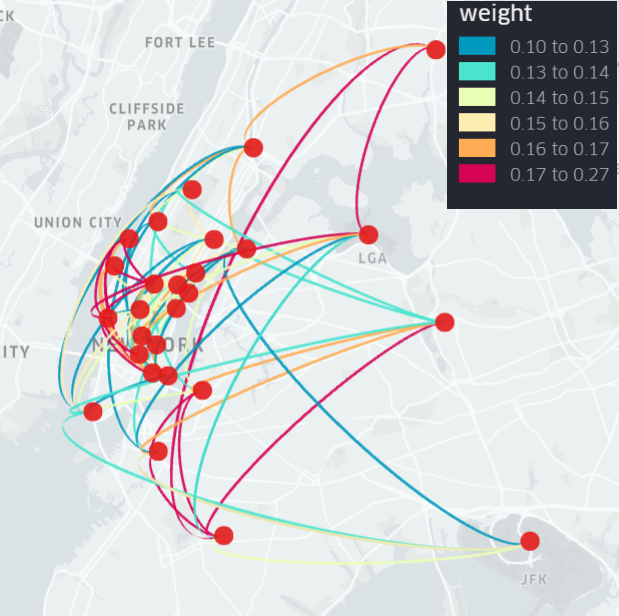}}
	   \hspace{18mm}
	  \subfloat[Ours: $\mathcal{G}_5$ (4 a.m.-5a.m.)]{
        \includegraphics[width=0.18\linewidth]{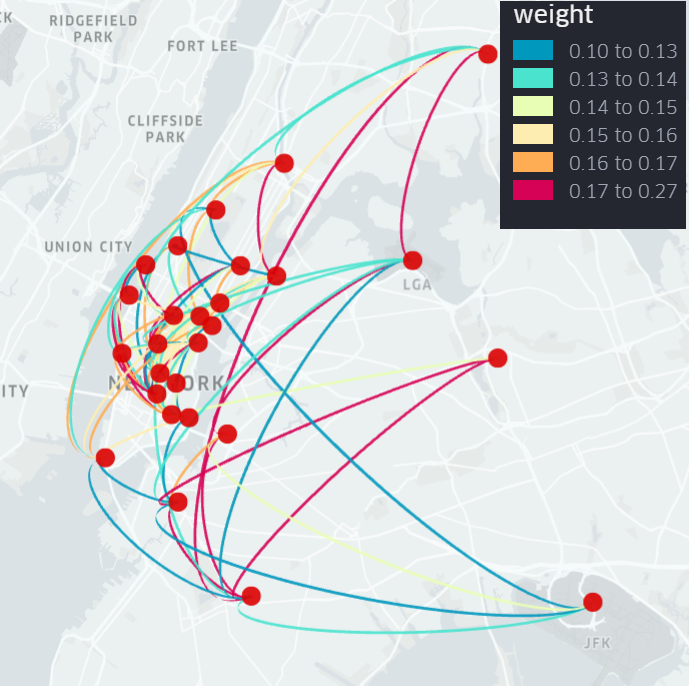}}
        \hspace{18mm}
      \subfloat[Ours: $\mathcal{G}_9$ (8 a.m.-9 a.m.)]{
        \includegraphics[width=0.18\linewidth]{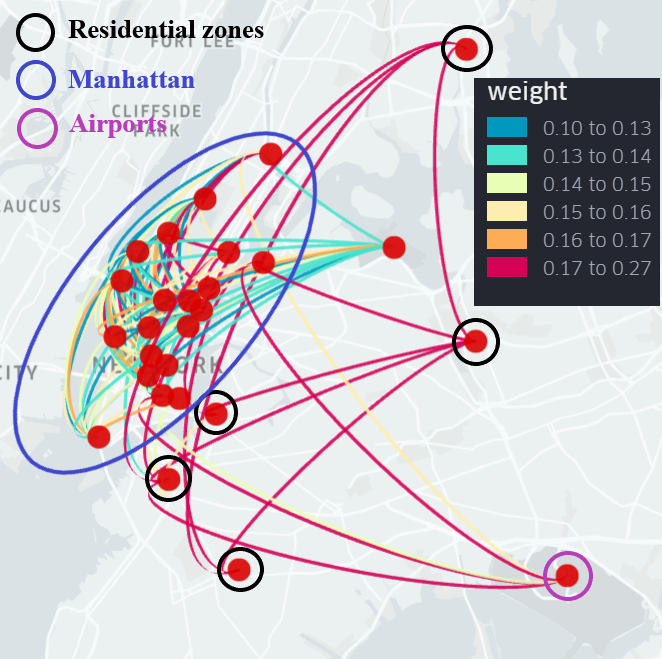}}

       \subfloat[TV-ho $\mathcal{G}_3$: (2 a.m.-3 a.m.)]{
        \includegraphics[width=0.18\linewidth]{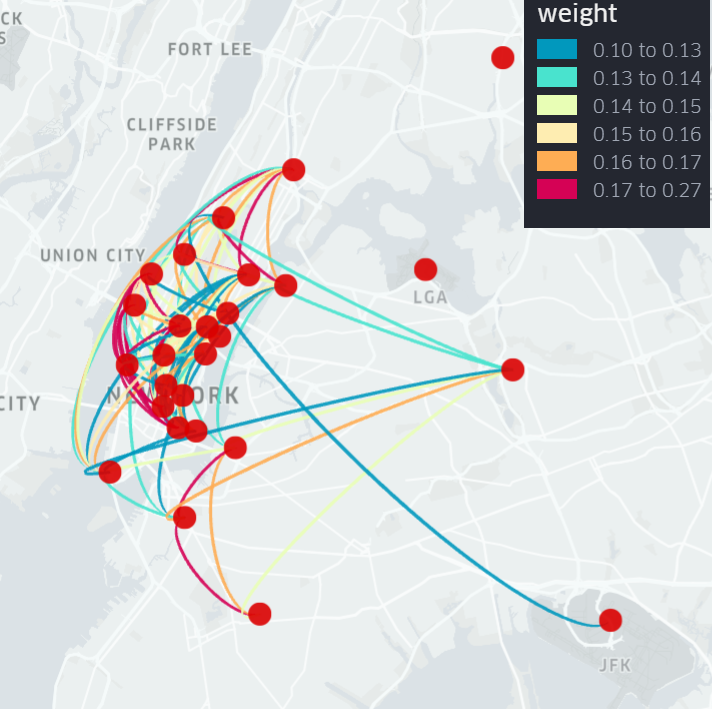}}
         \hspace{18mm}
        \subfloat[TV-ho: $\mathcal{G}_5$ (4 a.m.-5 a.m.)]{
        \includegraphics[width=0.18\linewidth]{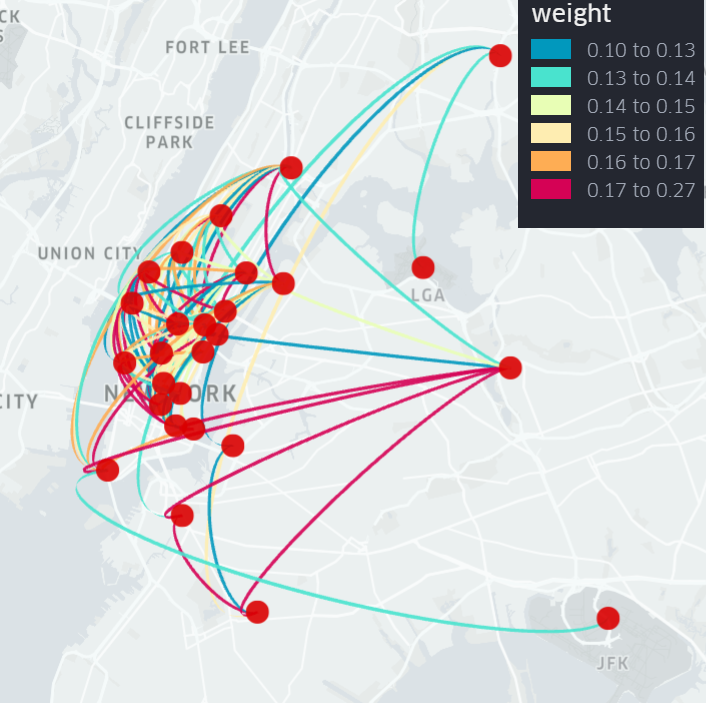}}
        \hspace{18mm}
        \subfloat[TV-ho: $\mathcal{G}_9$ (8 a.m.-9 a.m.)]{
        \includegraphics[width=0.18\linewidth]{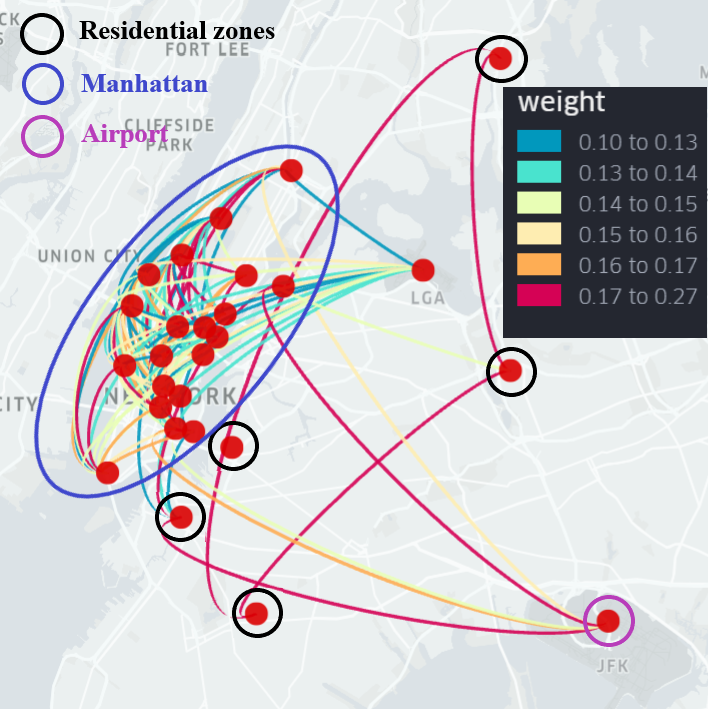}}
    	\caption{ The learned graphs of taxi zones of New York by our model and that of Model \eqref{classic time varying graph learning} (TV-ho)}
    	\label{fig-2BE6-2}
        \vspace{-1Em} 
\end{figure*}

\subsection{Real Data}
Our framework is also applied to Yellow Taxi Trip data of New York city\footnote{The data is available at https://data.cityofnewyork.us/Transportation/2018-Yellow-Taxi-Trip-Data.} to learn the time-varying travel relationships between different taxi zones. The data record timestamps and locations of pickups of taxi orders. We focus on data from 0 a.m. to 12 a.m. and learn a graph for each hour, which means that 12 time slots, as well as graphs, are finally obtained. The city are divided into 27 zones and the number of taxi pickups of each zones within 15 minutes  are taken as signals for that zone. A total of 80 graphs signals for each zone are collected, i.e., $\mathbf{X}_t\in\ \mathbb{R}^{27\times80}$ for each $t$, since we only select data of 20 workdays in September of 2018.

We then design a temporal structure $\mathcal{G}_N$, which is shown in Fig.\ref{fig-2BE6-1}, based on our prior knowledge of the variations of crowd flow networks. Note that temporal structure in Fig.\ref{fig-2BE6-1} is not a chain structure since the variations of crowd flow patterns at different time duration in one day are not uniform due to the diversity of travel behaviour. In the early morning, most people are in sleep and the crowd  mobility patterns may stay static. Therefore, we connect graphs in early morning with each other, i.e., from $\mathcal{G}_2$ to $\mathcal{G}_5$, even they are not adjacent in time. Additionally, it is common sense that crowd  mobility patterns changes significantly in rush hours and hence we set the smallest weight between $\mathcal{G}_7$ and $\mathcal{G}_8$.

We observe from  Fig.\ref{fig-2BE6-2} that $\mathcal{G}_3$ and  $\mathcal{G}_5$ learned by our method are similar despite they are not in consecutive time slots. It makes sense since the travel patterns in early morning should almost stay unchanged. However, temporal homogeneity assumption fails to capture this temporal characteristic. Compared with graphs of  $\mathcal{G}_3$ and $\mathcal{G}_5$, $\mathcal{G}_9$ learned by our method shows the following changes. 1) Connections between  residential zones are strengthened. This is caused by the fact that most people travel to work from home in rush hours. Therefore, the travel patterns of these zones are similar.  2) Connections between zones in Manhattan area are also strengthened due to the fact that more people take taxi to work area in Manhattan. However, these changes of TV-ho are less obvious than ours since it treats all variations equally.

\section{Conclusion}
\label{sec:conclusion}
In this paper,  we propose a general time-varying graph learning framework, under which temporal graph is employed to describe temporal structures. A distributed algorithm using ADMM framework is developed to solve the induced optimization problem. Experimental results show that our framework  outperforms the state-of-art methods when facing complicated temporal structures.


\end{document}